%% file: root.tex
\documentclass[letterpaper, 10 pt, conference]{ieeeconf}  

\IEEEoverridecommandlockouts                              

\usepackage{graphics} 
\usepackage{epsfig} 
\usepackage{amsmath} 
\usepackage{amssymb}  
\usepackage{bm} 
\usepackage[flushmargin,hang]{footmisc} 
\usepackage{flushend} 

\input{macros.tex}  
\usepackage{siunitx} 

\usepackage{pgfplots}
\pgfplotsset{compat=newest}
\usetikzlibrary{plotmarks}
\usetikzlibrary{arrows.meta}
\usepgfplotslibrary{patchplots}
\usepackage{grffile}
\pgfplotsset{plot coordinates/math parser=false}
\newlength\fwidth
\newlength\fheight

\usepackage{cite}

\title{\LARGE \bf
Risk-Aware Coverage Path Planning for Lunar Micro-Rovers Leveraging Global and Local Environmental Data}

\author{Shreya Santra$^{*}$, Kentaro Uno$^{*}$, Gen Kudo$^{*}$, and Kazuya Yoshida
    \thanks{
    S. Santra, K. Uno, G. Kudo, and K. Yoshida are with the Space Robotics Lab. (SRL) in Department of Aerospace Engineering, Graduate School of Engineering, Tohoku University, Sendai 980--8579, Japan. (E-mail: \tt{santra.shreya.a2@tohoku.ac.jp}) 
    }%
    \thanks{
    \textit{*These authors contributed equally to this work. The corresponding author is Shreya Santra.}
    }%
}%

\begin{document}

\maketitle
\thispagestyle{empty}
\pagestyle{empty}


\begin{abstract}

This paper presents a novel 3D myopic coverage path planning algorithm for lunar micro-rovers that can explore unknown environments with limited sensing and computational capabilities. The algorithm expands upon traditional non-graph path planning methods to accommodate the complexities of lunar terrain, utilizing global data with local topographic features into motion cost calculations. The algorithm also integrates localization and mapping to update the rover’s pose and map the environment. The resulting environment map's accuracy is evaluated and tested in a 3D simulator. Outdoor field tests were conducted to validate the algorithm's efficacy in sim-to-real scenarios. The results showed that the algorithm could achieve high coverage with low energy consumption and computational cost, while incrementally exploring the terrain and avoiding obstacles. This study contributes to the advancement of path planning methodologies for space exploration, paving the way for efficient, scalable and autonomous exploration of lunar environments by small rovers.

\end{abstract}


\section{INTRODUCTION}

Planetary rover missions have proved to be extremely invaluable for the exploration and survey over extended regions. Some of the most successful rover missions on the Moon include USSR's Lunakhod 2, CNSA's Chang'e4 Yutu-2, ISRO's Pragyan, and JAXA's SLIM LEV1/2~\cite{Lunar_missions}, and NASA's Opportunity, Curiosity and Perseverance achieved historical discoveries on the Mars~\cite{mars_missions}. Their ability to operate in remote and challenging environments have significantly advanced our understanding of the Moon and Mars.

In recent years, the space industry has taken a dramatic turn such that a considerable portion of its operation is supported by private entities. The major national space agencies are now partnering with the commercial sector with the aim to go back to the Moon for long-term exploration and utilization of lunar resources. This has opened up the possibility of using multiple smaller rovers that can significantly improve the performance of planetary exploration missions~\cite{Multi_robot}. This is specifically true when the region-of-interests are spatially distributed that requires dense scanning and sampling of the environment. The goal is that each rover explores the assigned area thoroughly, thus increasing the area of coverage efficiently. Coverage Path Planning (CPP) algorithms are often employed to address this, however, most of them assume a prior-knowledge of the environment~\cite{CPP_survey}. This presents a challenge when operating on extraterrestrial surfaces where the terrain feautures are not clearly defined. In such situations, the rovers must solely rely on their limited sensing capabilities and low-resolution orbiter data for navigation. 

\begin{figure}[t]
  \centering
  \includegraphics[width=\linewidth]{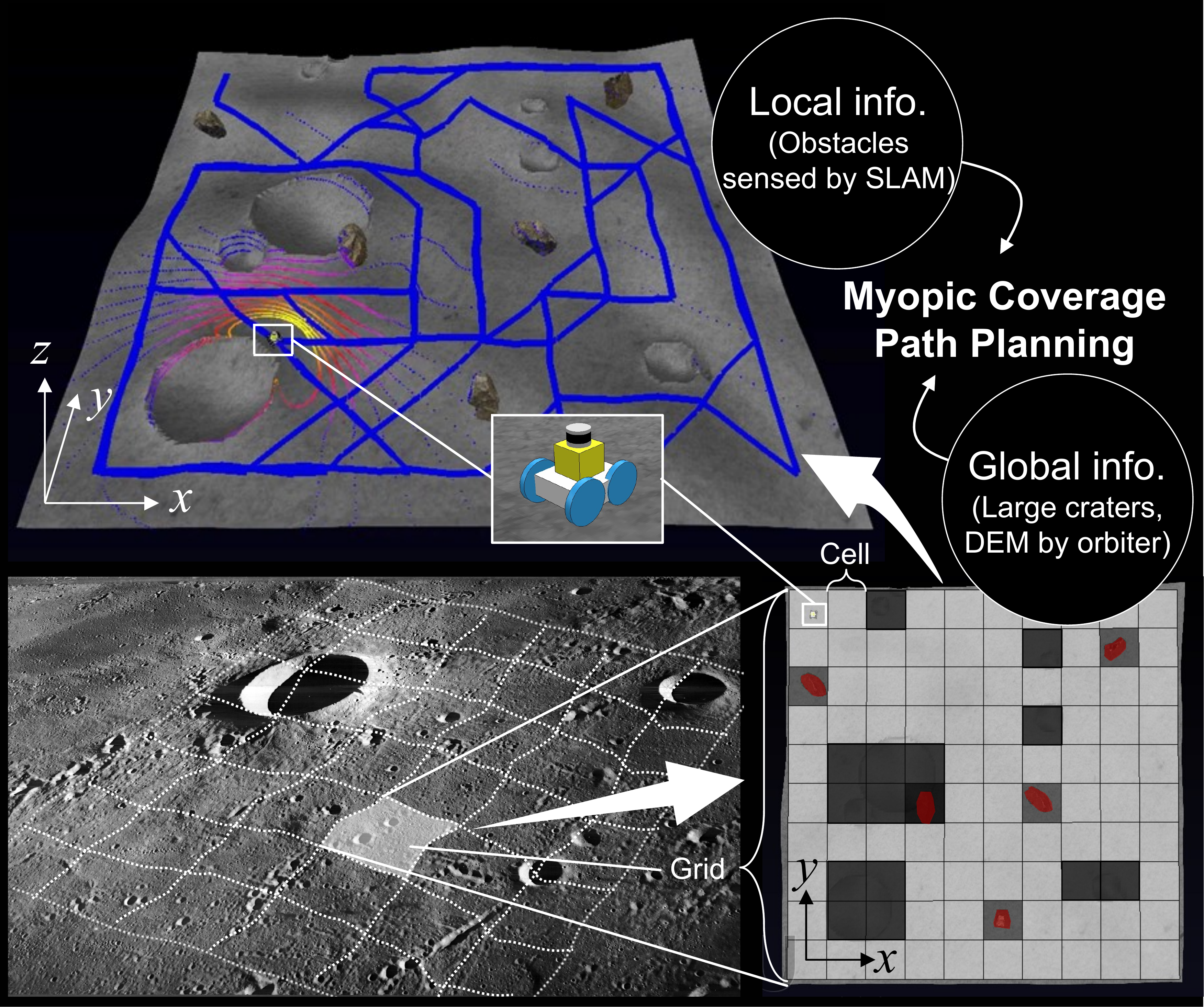}
  \caption{Concept of myopic coverage path planning by means of global and local level information for lunar rovers. A rover assigned per grid explores entirely to visit all the cell.}
  \label{paper_overview}
\end{figure}

This paper presents an approach towards coverage path planning using a single micro-rover in realistic 3D environment taking into account the constraints on the measurement range (\textit{myopic sensing}) and computation capability of the rovers. By combining the global information and the on-board sensor information as shown in~\fig{paper_overview}, it was demonstrated how a mobile rover is able to cover large exploration areas safely.

\section{RELATED WORKS}
\begin{figure}[t]
  \centering
    \begin{minipage}[t]{0.55\linewidth}
      \centering
      \includegraphics[width=1.0\linewidth,clip]{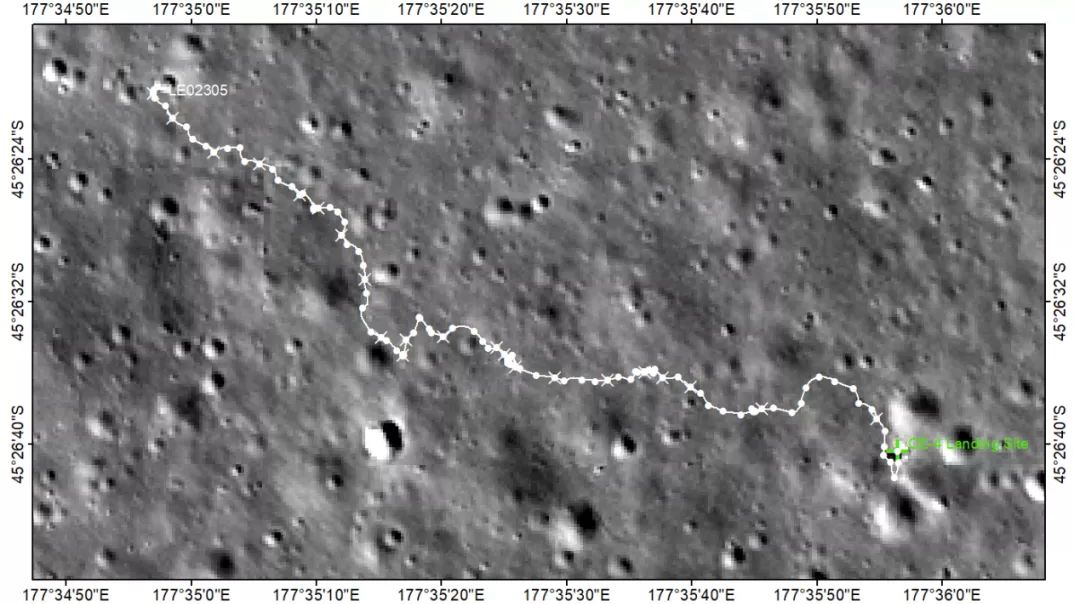}
      {\footnotesize (a) Global information based path planning}
    \end{minipage}
    \begin{minipage}[t]{0.4\linewidth}
      \centering
      \includegraphics[width=1.0\linewidth,clip]{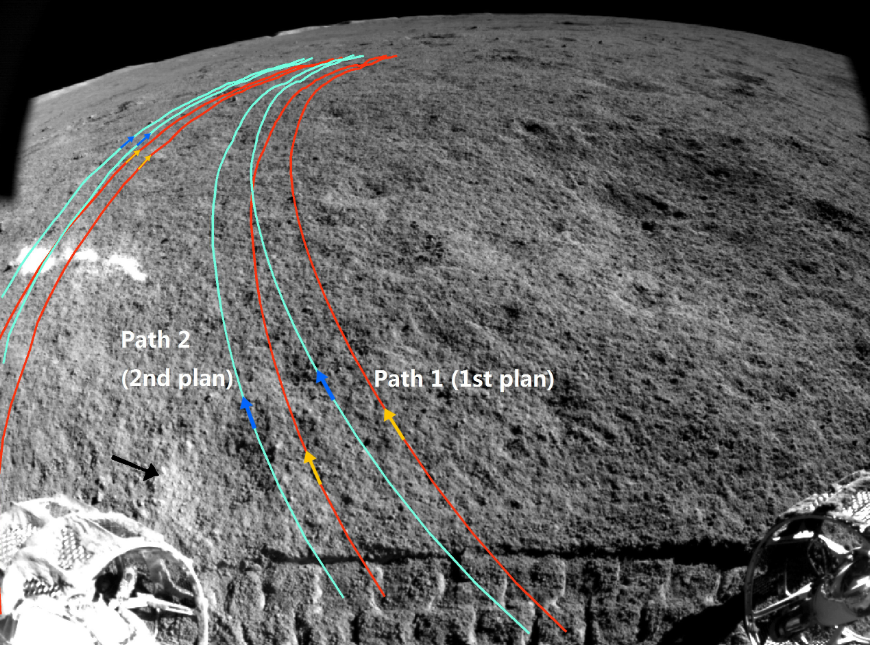}
      {\footnotesize (b) Sensor-based path planning}
    \end{minipage}
  \caption{Global and Local Path Planning of Chang'e 4 Yutu-2 rover ©CNSA}
  \label{Global_local_PP}
\end{figure}

In a known environment, with the presence of Global Positioning System (GPS) and global mapping, it is relatively simple to develop path planning and navigation solutions. In most cases, the area of interest is represented as a graph-structured map where obstacles and free areas are classified before the traversing path is programmed. In such cases, path planning is solved using various graph-based methods~\cite{raja2012optimal}. However, in reality, this type of information is not expected to be available in planetary exploration scenarios. 
In unfamiliar environments, the robot navigates, constructs a map, and moves toward the goal based on sensor input. Particularly for the robot driving in an unstructured environment, hybrid path planning has been adopted, considering the celestial surface as a known environment at the global level and an unknown environment at the local level~\cite{carsten2007global} (see~\fig{Global_local_PP}). Global path planning for rover missions is performed based on the pre-observed environmental information by the lunar orbiters and astronomically-known sunlight angle at the specific latitude of the Moon, which is taken into account to avoid the hazardous terrain (e.g. craters and steep hills)~\cite{oikawa2016thermal}.
On the other hand, various environmental features are only visible in-situ thus local path planning is essential. Furthermore, although the rover health-level index represented as power consumption, thermal balance and slippage can be pre-estimated from mathematical models~\cite{tanaka2024sensor}, these metrics should be monitored online with the rover's interoceptive sensors including visual, radio and wheel odometry to locally refine the path in order to maximize the mission safety. Particularly to predict traversability metrics, machine learning-based approaches has shown promising results under non-negligible uncertainty in nature ~\cite{skonieczny2019data,endo2022RAL}. 

Typical path planning problems set the starting position and the goal position waypoints, which does not consider the size of the traveled area, however in an exploration mission, the total coverage is a significant factor. CPP is a category of path planning designed to focus more on area coverage. Most CPP algorithms start their process with the decomposition of the environment. This process enables us to understand the environment and manage the state of exploration accurately. The most common decomposition method used is to discretize the environment into a grid~\cite{CPP_survey}. Then the goal of CPP is redefined as visiting all the small areas in this grid called a cell. Other methods rely on the position of obstacles or borders and the shape of the area decomposed to polygons. Coverage algorithms can be classified as heuristic or complete depending whether or not they guarantee complete coverage of the free space. Independently, they can be classified as either offline or online. This classification was originally proposed by Choset~\cite{Choset_CPP}. Offline algorithms assume that the environment is known in advance and stationery. On the contrary, online algorithms do not assume full prior knowledge of the environment to be covered and utilize real-time measurement to sweep the target space. Therefore, it is also called sensor-based coverage path planning.

General applications of CPP can be found in floor-cleaning robots, lawn mowers, automated harvesters, structural inspection and surveillance ~\cite{CPP_survey}. Since accurate information about the environment is accessible both in outdoor and indoor environments, most algorithms rely on complete knowledge of the environment. One of the well-studied branches of CPP is Spanning Tree Coverage (STC)~\cite{STC}. It is a group of graph-based approaches, which guarantee the fastest coverage by providing a single path. STC constructs a spanning tree without any loops throughout the environment, and then the robot circumnavigates to complete the coverage. It can be used online for unknown environments, but it requires recalculation of the path each time an obstacle appears. In other words, the most efficient path will be obtained only after it visits all the cells.

To summarize, there are several CPP algorithms that proved to be efficient  for terrestrial applications, cannot be used for the unknown planetary surfaces or is not suitable for micro-rovers due to computational complexity necessary for localization and navigation. Even though randomization is the best choice to reduce computational load, it is impractical to cover the entire area within a given time in most scenarios.

``Myopic'' refers to the shortsightedness of microrovers specifically due to the limited height of mounted sensors. To address these limitations, a simple and lightweight Myopic Path Planning (MPP) method is devised. It differs from most of the conventional path planning methods as prior knowledge is not necessary. The rovers incrementally proceed with exploration by following this algorithm and achieve complete coverage in a given time. The advantage of this algorithm is that it is completely distinct from a randomized algorithm, has higher efficiency, and scalable to any number of rovers or size of exploration area~\cite{Laine2021FSR}.


\section{PROBLEM FORMULATION}
MPP follows the fundamentals of CPP, i.e., decomposition of the environment into grids of evenly sized finite set of cells to explore an environment in its entirety. In this work, the region is discretized into a grid with the total number of cells represented by the value $n = \textit{rows} \times \textit{cols}$. The four different states of each cell is defined as follows:
\begin{itemize}
 \item \textit{Unknown}: Initial state of all cells, until visited by the rover
 \item \textit{Free}: The detected cells after being sensed by the rover.
 \item \textit{Obstacle}: The sensed cell contains obstacles and cannot be visited by the rover.
 \item \textit{Visited}: The cells that the rover has already sensed and visited at least once.
\end{itemize}

\begin{figure}[t]
  \centering
  \includegraphics[width=0.7\linewidth]{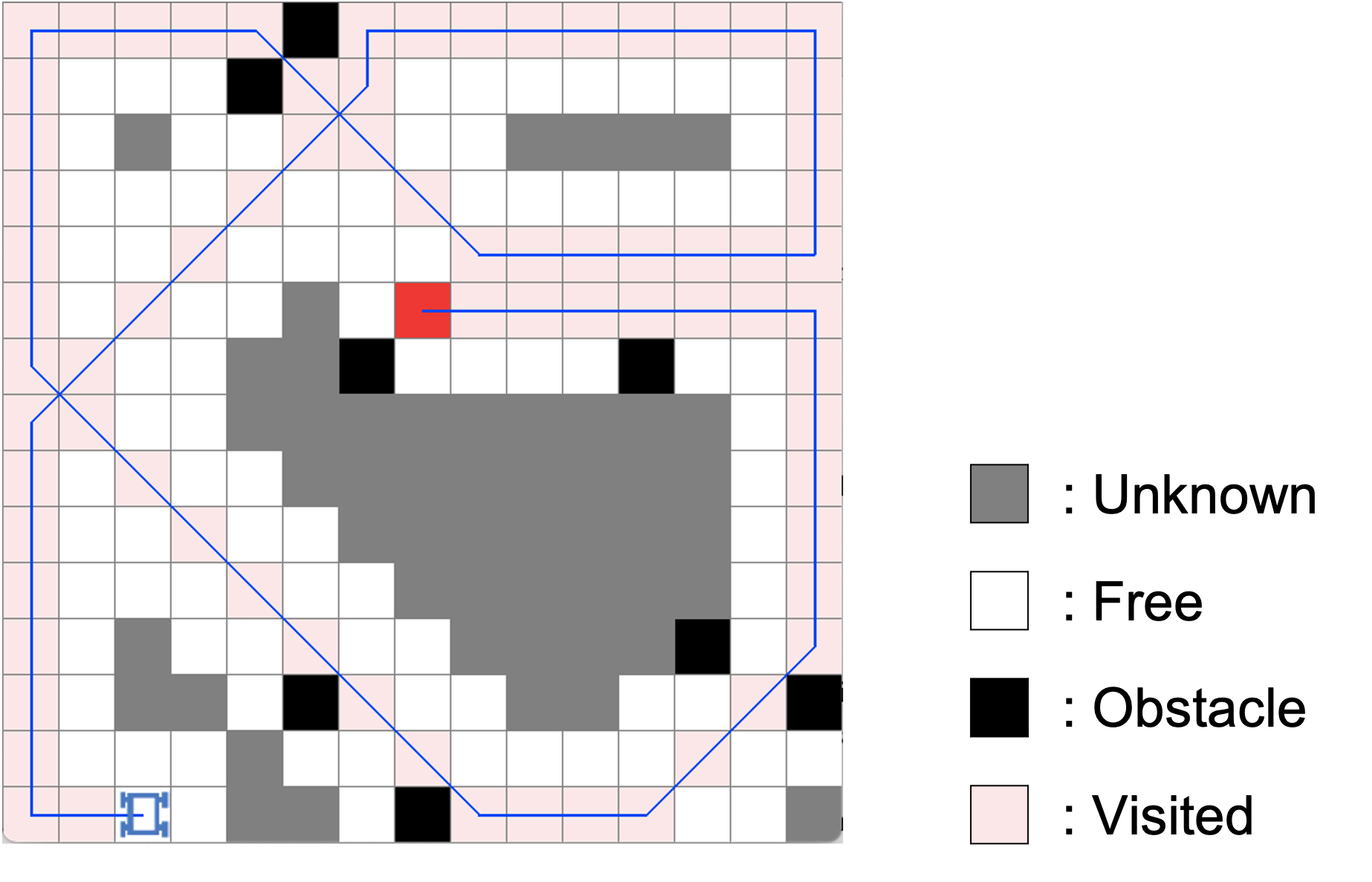}
  \caption{Grid representation of exploration area.}
  \label{grid_representation}
\end{figure}

At the beginning of the exploration, all the cells are registered as unknown. The goal of the exploration is determined as visiting as many free cells as possible except for obstacles under a given time. The onboard sensors are designed for navigation or scientific mission and selected to maximize the rover’s capability to localize itself in the exploration area. Traditionally, for planetary rovers, vision sensors such as stereo and RGBD cameras are most popularly used to sense the surroundings, however, they have limited field of view and sensing range, which lowers the exploration efficiency. In addition, dark environments make it difficult to extract feature points, which can lead to less accurate localization and loss of odometry in worst cases. For dense coverage, this disadvantage is too critical to dismiss~\cite{LiDAR_2021}. For these reasons, LiDAR was chosen as the sensor in this study. 
The grid representation of exploration area is shown in~\fig{grid_representation}. The size of the environment is \textit{\SI{60}{m} × \SI{60}{m}} containing randomly distributed obstacles, which amount to 5$\%$ of the total cell count. Each cell size is assumed to be \textit{\SI{1}{m} × \SI{1}{m}} accounting for the LiDAR's ability to detect any terrain variability within this range. 

\subsection{Coverage Cost Function}

The rover explores the region by visiting the cells one by one, iterating the same process in each cell. At every iteration, the rover acquires information of its surroundings as a first step. With the LiDAR onboard, the rover can sense the eight neighboring cells around it. From its pose on a map, the surrounding cells can be updated to be  free, obstacle, or visited. Based on this information and the rover’s orientation, the cost function assigns motion cost to each neighboring cell. As a final step, the rover selects the cell with minimum cost and moves towards it. The position of the rover is updated and the movement from 
\((x_{i},y_{i})\) to \((x_{i+1},y_{i+1})\) determines the new orientation of the rover. The basic cost function that defines the motion of the rover is represented by the function: 
\begin{equation}
\label{basic_cost_func}
mc = mc_{\text{static}} + mc_{\text{visited}}V_{i}
\end{equation}
where, first component, \(mc_{\text{static}}\) is a basic motion cost assigned to surrounding cells based on the orientation of the rover. The second component, \(mc_{\text{visited}}\), is a penalty for visited cells that are added to the cost based on relative direction of the rover's movement. This value should be greater or equal to 1.1 to prioritize non-visited cells. \(V_{i}\) represents the number of times the rover visits the cell. Without \(mc_{\text{visited}}\), the rover is often stuck in an infinite loop. This cost becomes more obvious as the environment becomes larger and revisitation is avoided, as it works as a natural effect pushing the rover away from the highly visited areas towards the not visited region.

\subsection{Digital Elevation Model}
\begin{figure}[t]
  \centering
    \includegraphics[width=.7\linewidth]{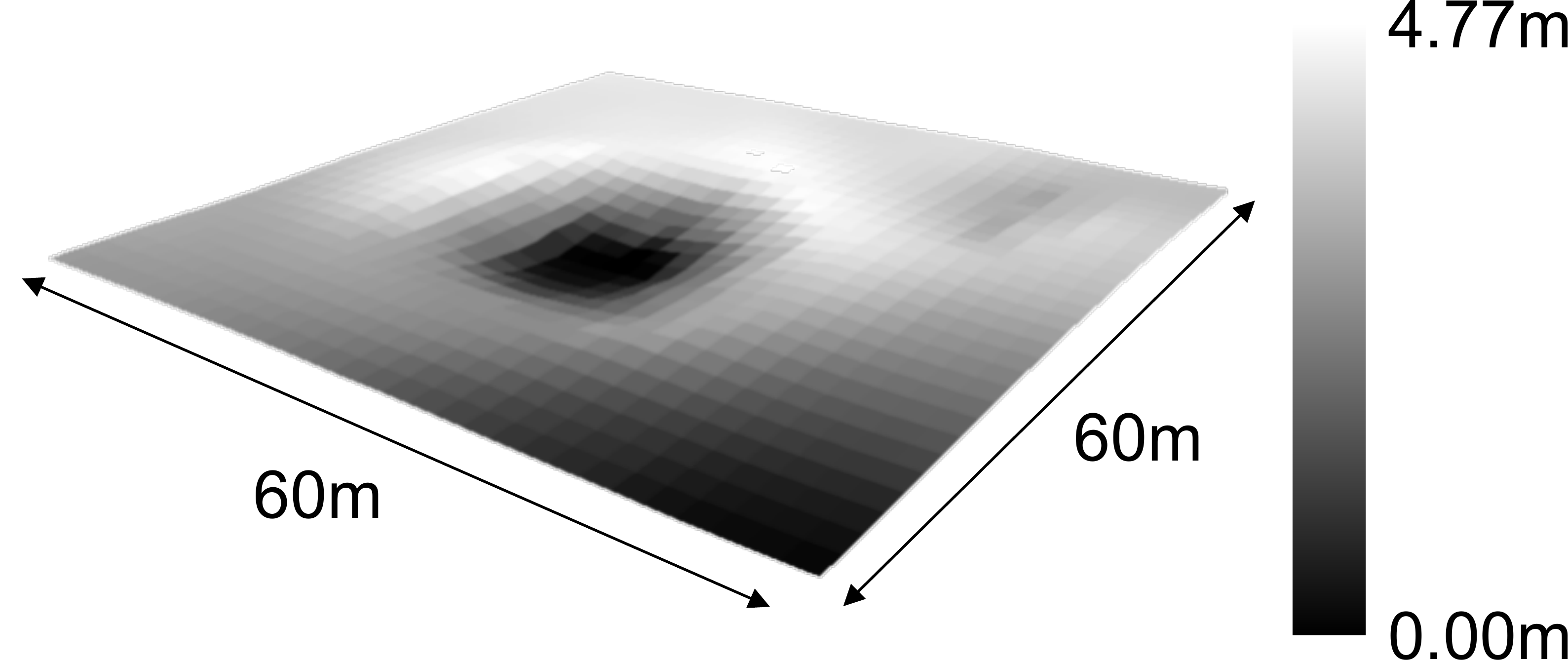}\\
    {\footnotesize (a) 3D view of DEM near Apollo14 landing site.} \\
    \vspace{3mm}
    \begin{minipage}[t]{0.48\linewidth}
      \centering
      \includegraphics[width=.9\linewidth,clip]{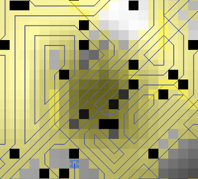}
      {\footnotesize (b) $\beta = 0$.}
    \end{minipage}
    \begin{minipage}[t]{0.48\linewidth}
      \centering
      \includegraphics[width=.9\linewidth,clip]{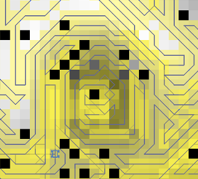}
      {\footnotesize (c) $\beta = 0.7$.}
    \end{minipage}
  \caption{CPP results by changing the weight of cost function: with ($\beta = 0.7$) and without ($\beta = 0$) DEM based cost term.}
  \label{DEM_cost}
\end{figure}

The digital elevation model (DEM) is the digital representation of elevation data of planetary surfaces, generated from satellite data. As for the lunar surface, high-resolution DEM data is provided by NASA, openly accessible at the NASA Planetary Data System (PDS)~\cite{pds}.
~\fig{DEM_cost}(a) shows the DEM of the area near Apollo14 mission where highlands and lowlands can be visually identified. It is useful to understand the topography of the lunar surface and is utilized for simulation.

To take terrain geometry into account, another cost based on DEM is added to the cost function. This cost can be calculated in advance while myopic cost is calculated in real-time based on the rover’s sensor data. Thus it is regarded as hybrid path planning taking into account both global and local information. The difference in height between two cells can be translated into the slope gradient. Based on this assumption, an additional cost is introduced to consider the absolute difference in height between adjacent cells:
\begin{equation}
\label{gradient_slope_formula}
mc_{\text{DEM}}(p_{t+1}) =|h(p_{t+1)} - h(p_{t)}|
\end{equation}

The new cost function is defined as the sum of the two elements, the basic motion cost and the DEM-based cost:
\begin{equation}
\label{DEM_cost_func}
mc = \alpha (mc_{\text{static}} + mc_{\text{visited}}V_{i}) + \beta mc_{\text{DEM}}
\end{equation}

The weights $\alpha$ and $\beta$ are such that $\alpha + \beta = 1$. Three metrics are introduced to evaluate the performance of the algorithm from the new cost function: the path length ratio, energy consumption, and maximum pitch angle of the rover. The efficiency of the path planning algorithm can be evaluated by the path length required to accomplish a specific percentage of coverage. As a reasonable compromise, the coverage is set to 95$\%$ completeness for this simulation. The reason is that it requires 15$\%$ more path length to visit the remaining 5$\%$ cell in the final phases of the exploration since the algorithm does not actively use sensing memory to locate the final cells~\cite{Laine2021FSR}.The ratio of path length to total cell count, defined as path length ratio, is introduced as
\begin{equation}
\label{Path_lenth_ratio}
\textit{Path Length Ratio} = \frac{\textit{Total Path Length}}{\textit{Total number of cells}}
\end{equation}

The exploration is more efficient if this ratio is minimized, irrespective of the target coverage. For a single rover exploration, an optimal result would be to explore all cells atleast once yielding a ratio of 1.0. 

\subsection{Energy Consumption}

Energy consumption, as indicated by the term, is the overall electricity usage required to accomplish the exploration task, and constitutes a pivotal consideration. The energy required to move from one cell to another is affected by the distance between the cells and the slope gradient. In this work the energy consumption is formulated based on the work by Oikawa et al.~\cite{oikawa2016thermal}. Regarding the rover's localized movement in sandy terrain without a steering system, an efficient locomotion strategy, focused on minimizing power consumption, involves a combination of longitudinal motion and spot turns at each step, given by: 
\begin{equation}
\label{energy_consumption}
E_{\text{total}} = E_{\text{forward}}+ E_{\text{rotate}}
\end{equation}

where \(E_{\text{forward}}\) is the total energy for longitudinal motion and \(E_{\text{rotate}}\) is the total energy for spot turn. If there is a considerable difference between the height of the adjacent cells, it can be regarded as a hill or steep slope. This type of terrain requires more energy to traverse and potentially causes slip. 


In this paper, the rover’s pitch angle is used for risk assessment. Slip ratio, which represents the state of traversability, increases exponentially as the pitch angle increases and velocity decreases. The value of the pitch angle should be maintained low to ensure less risk.
By changing the value of the weights $\alpha$ and $\beta$ different path lengths, energy consumption, and maximum pitch angle are obtained. As shown in \fig{DEM_cost}(b) for $\beta=0$ the rover tried to cross the crater because 3D terrain is not considered. This indicates that the rover selects a path without recognizing the existence of the crater. On the other hand, when DEM-based cost is incorporated, the rover traveled along the crater's circumference as seen in \fig{DEM_cost}(c). Therefore, for smaller $\beta$ most of the energy is consumed in moving up a steep slope while for larger $\beta$ it is consumed due to a longer path. This shows that the optimal trajectory can be obtained by modifying the parameters of the cost function even if the area contains unknown obstacles.

\section{SIMULATION EVALUATION}
A 3D simulator environment was developed in CoppeliaSim~\cite{coppelia} using the Bullet physics engine library to simulate and evaluate the proposed algorithm. 

\subsection{Modelling of Rover Testbed}
CLOVER is a small rover platform developed at the Space Robotics Lab (SRL) for research purposes. It is a four-wheeled rover composed of aluminum frame and based on Robot Operating System (ROS1)~\cite{ros}. All wheels are actuated by independent motors. It has a simple structure with dimensions \SI{40}{cm} × \SI{35}{cm} × \SI{30}{cm} equipped and weighing about \SI{7}{kg}. It has onboard computational devices and Velodyne's 3D LiDAR VLP-16 mounted on it. 
The rover was modeled as an assembly of some basic structures such as a cylinder and a cuboid with a LiDAR module. The path planning scripts are implemented external to the simulator, communicating through the ROS architecture, interacting with the wheels and sensors, and performing low-level computations.

\subsection{Modelling of Lunar Terrain}
The Apollo 14 DEM maps were pre-processed using Geospatial Data Abstraction Library (GDAL)~\cite{USGS} and converted to a high-resolution lunar terrain on Unity 3D~\cite{masawat_lunar}. This was then imported to the simulation scene. Various sizes of craters, gentle hills and rocks was also placed manually. Being a raster data, the terrain is represented as consecutive cells with each cell containing the height information. It is highly compatible with the CPP, whose constant is the decomposition of the environment into cells. 

\subsection{Integration of Bug Algorithm}
\begin{figure}[b]
  \centering
  \includegraphics[width=\linewidth]{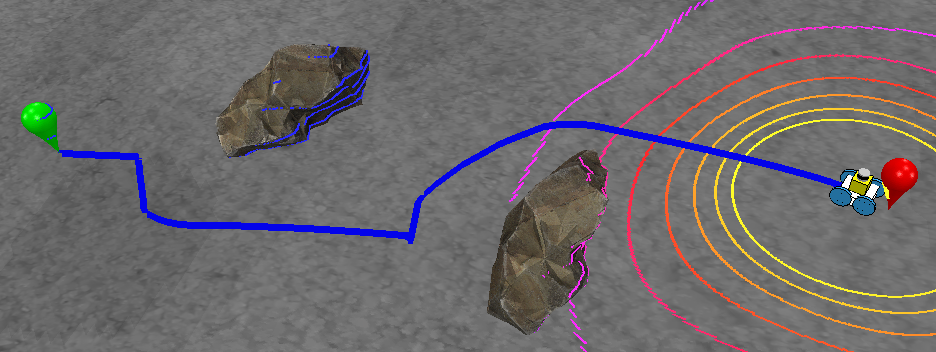}
  \caption{Myopic path planning using bug algorithm and SLAM in the simulator.}
  \label{bug_algo}
\end{figure}

A reactive path planning solution, called the bug algorithm was integrated with myopic path planning, to avoid obstacles during inter-cell movement. In a completely known environment,it is easy for robots to create the binary map and use graph-based algorithms. However, in a partially known environment, the functions of navigation and obstacle avoidance become necessary to search a safe path. The bug algorithm can form the optimal path to the target and take a quick decision to avoid the obstacles by using sensorial information~\cite{bug_algo_new}. Depending on the relative position of the obstacles detected from the point clouds obtained from LiDAR, the rover will travel clockwise or counterclockwise around it to avoid it. The rover sets an artificial boundary at a certain distance from the obstacle and travels in such a way that it follows the boundary.~\fig{bug_algo} shows this algorithm in action in simulator. 


\subsection{SLAM Implementation}
Considering the limitations of autonomous exploration by micro rovers, the Simultaneous Localization and Mapping (SLAM) algorithm must be capable of accurate and real-time localization in feature-less environments. Large-scale exploration is possible with a graph-based SLAM framework~\cite{graph_SLAM}. The reason for this is the ability to detect loop closure, which is a method of optimizing the rover’s pose and map by revisiting a previously visited location thus minimizing the accumulated errors. After an extensive survey and trade-offs, HDL Graph SLAM developed by Koide et al.~\cite{HDL_SLAM} was selected for this study due to its adaptability to the challenging environment of planetary exploration. Its ROS package enables easy installation and is compatible with CLOVER and CoppeliaSim. The point cloud data is fed to the HDL Graph SLAM as a rostopic, and output position and orientation are returned to the simulator. The accuracy of localization is evaluated by the Mean Absolute Error (MAE) defined as:
\begin{equation}
\label{Lidar_map_transformation}
M\!AE = \sum_{i=1}^{N} |x_{1i} - x_{2i}| 
\end{equation}
where $x_{1i}$ is a position estimated by SLAM and $x_{2i}$ is the ground truth. To obtain accurate MAE, the x-axis of LiDAR and the map should be aligned. Comparing the actual terrain and the obtained map, a sufficiently accurate map was generated that was dense enough for offline path planning, construction planning of structures on the Moon, and for assessment of landing points for the next mission. Thus, the path planning algorithm was extended to a 3D environment and full autonomous exploration was achieved by integrating SLAM. 


\section{SIM-TO-REAL EXPERIMENTS}
\begin{figure}[t]
  \centering
  \includegraphics[width=1.0\linewidth]{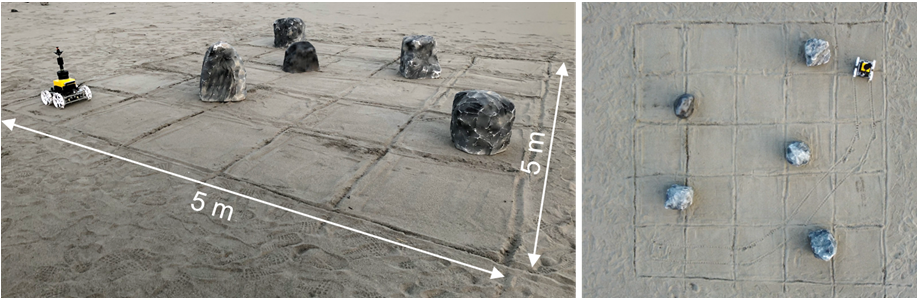}
  \caption{The field test site on the beach sandy soil with a grid.}
  \label{beach_site_grid}
\end{figure}

To validate the proposed algorithm, field tests were conducted in real-world scenarios. The algorithm was tested on the CLOVER micro-rover to evaluate its applicability in outdoor environments, on clayey and sandy soil. The similarity to the lunar surface was a focused point when selecting the test sites. The experiment area is a \SI{5}{m} × \SI{5}{m} field divided into grids, each cell of \SI{1}{m} × \SI{1}{m}. Fake styrofoam rocks are placed as obstacles in the area. The placement of the rocks was chosen randomly. The initial position and posture of the rover are shown in the~\fig{beach_site_grid}. 
The ground truth of the rover's 3D position is measured using a total station. 
For real-time path planning in outdoor environment, the experiment was conducted in the following steps:
\begin{itemize}
 \item Pre-scan: The objective is to collect point cloud data from LiDAR in Outdoor and reconstruct 3D terrain.
 \item Offline path planning: The reconstructed 3D terrain is imported into the simulation, and path planning is performed using the robot’s onboard computer.
 \item Real-time path planning: After validation in simulation, perform real-time path planning and fully autonomous exploration outdoors.
\end{itemize}

\subsection{Outdoor Tests Results}

\subsubsection{Terrain Reconstruction and Off-line Path Planning}

\begin{figure}[b]
  \centering
  \includegraphics[width=\linewidth]{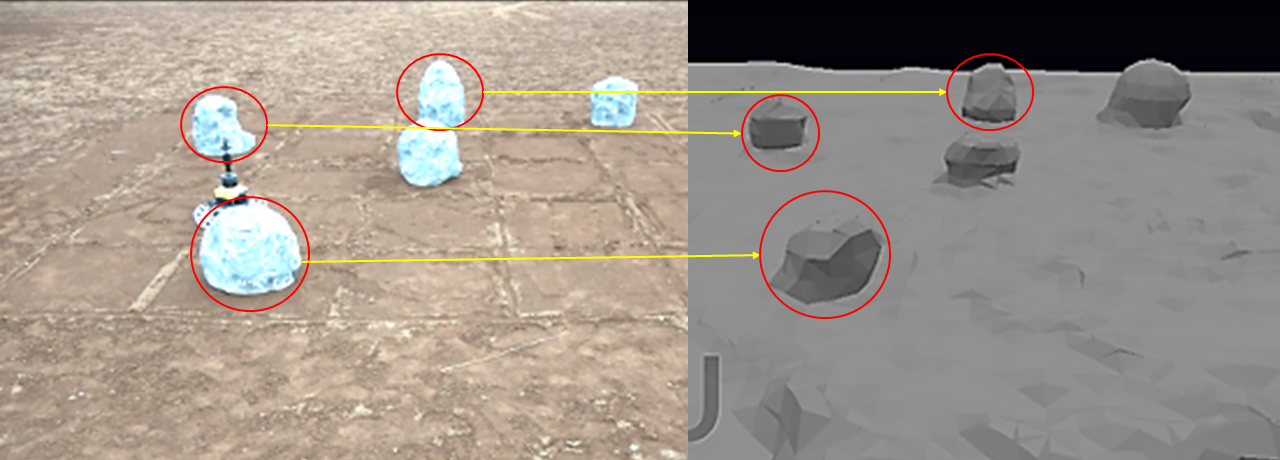}
  \caption{Reconstructed terrain of clayey soil test site in simulator.}
  \label{terrain_reconstruction}
\end{figure}

~\fig{terrain_reconstruction} shows the reconstructed terrain of the test site with clayey soil obtained by scanning with a LiDAR.
The off-line path planning was performed on the reconstructed terrain. The trajectory of the rover generated in the simulation proved the applicability of myopic path planning for coverage missions in real-world scenarios.

\subsubsection{Real-Time Path Planning}
\begin{figure}[t]
  \centering
  \includegraphics[width=0.7\linewidth]{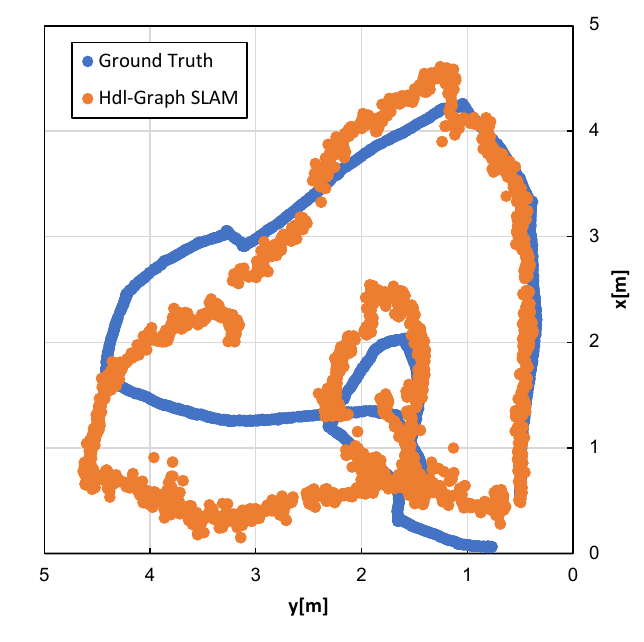}
  \caption{Accuracy of localization at the flat ground with clayey soil.}
  \label{Localization}
\end{figure}

Finally, the path planning was performed at the test sites by first scanning the exploration areas. The results of comparison with ground truth is presented in~\fig{Localization}.

The performance is evaluated based on the observation and measured value summarized in Table \ref{tab:Effiency_coverage}. A pre-scan to identify the gaps between the simulation and the real environment resulted in an accurate map of the field, which was then imported into the simulation environment for offline path planning. The optimization with loop closure significantly contributed to achieving higher coverage. Overall, the results of several experiments, both indoor and outdoor, demonstrated that MPP in the simulation can be applied in the real environment with simple adjustments.

\begin{table}
\centering
\caption{Efficiency and Energy consumption obtained from field tests}
\begin{tabular}{|c | c | c|} 
 \hline
 \textbf{Test Site} & \textbf{Coverage [$\%$]} & \textbf{MAE [m]} \\ 
 \hline\hline
 Outdoor field with clayey soil & 80 & 0.41 \\ 
 Outdoor field with sandy soil & 65 & 0.32 \\
 \hline
\end{tabular}
\label{tab:Effiency_coverage}
\end{table}

\section{DISCUSSION}
This coverage result was achieved from three distinctive works. The first was the expansion of the 2D myopic path planning algorithm to the 3D environment, which enables us to consider the lunar terrain. The proposed method introduces DEM as local topography to take into account the motion cost for the rover. The algorithm was implemented in the rover in a 3D simulator. A map of the environment was obtained as the final output of SLAM and its accuracy was qualitatively discussed. Field tests proved the significance of our algorithm in real-world scenarios. The only prior information required is the rough terrain DEM, and other sensing-based path planning is performed onboard in real-time. This path planning algorithm, designed for small rovers, can achieve high coverage with minimal energy consumption in a given time in environments with unknown obstacles. During the outdoor tests, some failure cases were observed due to wheel slippage and localization errors. The need for loop closure, as indicated by the field tests, raises the possibility of further development of this algorithm.

\section{CONCLUSIONS}
Robotic autonomy is considered an extremely important and fundamental requirement for future  lunar missions for resource utilization and future settlements. The fields related to mobile robotics and automation, such as path planning and SLAM, are booming. However, conventional path planners used in space exploration assume simple movement from one point to another with limited information at the microscopic level, which is inadequate for resource prospecting missions. This paper demonstrates a solution which can explore an area with a rover of limited sensing range, through realistic simulations and validation experiments. Furthermore, the algorithm is light-weight, scalable, and does not depend on the shape of the rover or the exploration area. It can be applied to exploration by multiple rovers by combining it with a high-level mission planner. Additional considerations such as costs based on terramechanics, communication between rovers~\cite{Comms}, etc. can be integrated to achieve more realistic path planning and are possible next steps for multi-robot large-scale exploration.

\section*{Acknowledgement}
The authors would like to thank Kenta Sawa, Masahiro Uda, Ayumi Umemura, Gabin Paillet and Mickael Laîné for their invaluable discussion as well as great support in the robot development and the field testing.



\bibliography{reference.bib}

\end{document}

%% file: macros.tex
\newcommand{\fig}[1]{Fig.~\ref{#1}}

\def\epsgaiji#1{\leavevmode\kern-0.025zw\raise-.37zh\hbox{%
  \epsfile{file=#1,width=1.05zw}}\kern-0.025zw}
\newcommand{\MARU}[1]{{\ooalign{\hfil#1\/\hfil\crcr\raise.167ex\hbox{\mathhexbox20D}}}}

\usepackage{array}
